\documentclass{article}

\usepackage{times}
\usepackage{graphicx} 
\usepackage{subfigure} 
\usepackage{amsmath}
\usepackage{natbib}
\usepackage{algorithm}
\usepackage{algorithmic}
\usepackage{hyperref}
\usepackage{array}
\usepackage{mathtools}

\usepackage[accepted]{icml2015} 

\newcolumntype{L}[1]{>{\raggedright\let\newline\\\arraybackslash\hspace{0pt}}m{#1}}
\newcolumntype{C}[1]{>{\centering\let\newline\\\arraybackslash\hspace{0pt}}m{#1}}
\newcolumntype{R}[1]{>{\raggedleft\let\newline\\\arraybackslash\hspace{0pt}}m{#1}}
\newcommand\myeq{\stackrel{\mathclap{\normalfont\small\mbox{def}}}{=}}

\icmltitlerunning{Comparative Study on Generative Adversarial Networks}

\begin{document} 

\twocolumn[
\icmltitle{Comparative Study on Generative Adversarial Networks}

\icmlauthor{Saifuddin Hitawala}{saifuddin.hitawala@uwaterloo.ca}
\icmladdress{David R. Cheriton School of Computer Science, University of Waterloo}

\icmlkeywords{machine learning, generative models, literature review, adversarial networks}

\vskip 0.3in
]

\begin{abstract} 
In recent years, there have been tremendous advancements in the field of machine learning. These advancements have been made through both academic as well as industrial research. Lately, a fair amount of research has been dedicated to the usage of generative models in the field of computer vision and image classification. These generative models have been popularized through a new framework called Generative Adversarial Networks. Moreover, many modified versions of this framework have been proposed in the last two years. We study the original model proposed by Goodfellow et al. \yrcite{goodfellow2014generative} as well as modifications over the original model and provide a comparative analysis of these models.
\end{abstract} 

\section{Introduction}
\label{introduction}

Machine learning as a field has grown rapidly in the past decade. Fields ranging from banking to healthcare, from marketing to autonomous vehicles, all make use of machine learning techniques. Accordingly, tremendous amount of research in both academia and industry is dedicated to the efficient use of techniques and methodologies and development of new techniques using machine learning. Research papers related to natural language processing, sentiment analysis, computer vision and object recognition, recommender systems and information retrieval are published almost every day. Specifically, in the field of computer vision and image classification, a plethora of research has been conducted in recent years. Some of this work is related to generative models and their usage in the fields of supervised, semi-supervised and unsupervised learning.

Generative models as compared to discriminative models have had less of an impact, due to the difficulty of approximating many intractable probabilistic computations arising in maximum likelihood strategies and due to the difficulty of leveraging piecewise linear units  in generative context. Goodfellow et al. \yrcite{goodfellow2014generative} overcame this problem by proposing a framework called adversarial nets, where a generative model is pitted against an adversary which is a discriminative network and learns to determine whether a sample came from the generator or from the training data. This model used multilayer perceptrons and was successful at generating samples similar to the MNIST dataset images. Consequently, due to the success and popularity of adversarial nets, many modifications over the original model have been proposed. Some of the modified versions of Generative Adversarial Nets (GANs) that we review in this study are: Conditional Generative Adversarial Networks (CGAN) \cite{mirza2014conditional}, Laplacian Pyramid of Adversarial Networks (LAPGAN) \cite{denton2015deep}, Deep Convolutional Generative Adversarial Networks (DCGAN) \cite{radford2015unsupervised}, Generative Recurrent Adversarial Networks (GRAN) \cite{im2016generating}, Adversarial Autoencoders (AAE) \cite{makhzani2015adversarial}, Information Maximizing Generative Adversarial Networks (InfoGAN) \cite{chen2016infogan} and Bidirectional Generative Adversarial Networks (BiGAN) \cite{donahue2016adversarial}.

\subsection{Motivation}
\label{motivation}

The biggest motivation behind this study was the absence of any other survey performing a comparative analysis of the  different versions of Generative Adversarial Networks (GANs). This absence was supplemented by the popularity of generative models in recent years and their extensive areas of application in the fields of computer vision and image classification. This comparative study tries to compare the original and modified versions of GANs and compare them on the basis of their network architectures, basic methodology and technique behind modifications, gradient optimization calculations, experimental performance and applications. 

\subsection{Paper Structure}
\label{structure}
The paper is structured in different sections. In Section 2, a background on Generative Adversarial Networks is provided. Section 3 talks about the different versions of GANs, their methodologies, model architectures, and experimental performances. In Section 4, a comparison between different versions of GANs is provided based on different parameters. Finally, Section 5 concludes the paper suggesting future work in the field of generative models.  

\section{Background: Generative Adversarial Networks}
\label{background}

Generative Adversarial Networks \cite{goodfellow2014generative} consist of a pair of models called the generator and discriminator. The generative models can be thought of as a group of thieves trying to generate counterfeit currency whereas the discriminative model can be thought of as police trying to detect the counterfeit currency. Thus, the entire framework resembles a two-player minimax game where the generator tries minimize its objective function and the discriminator tries to maximize its objective function. The objective of this game is given as follows:

\begin{multline}
\min_{G} \max_{D} V(D, G) = E_{x \sim p_{data}(x)}[logD(x)] \\ + E_{z \sim p_{z}(z)}[log(1 - D(G(z)))]
\end{multline}

\begin{figure}[ht]
\vskip 0.2in
\begin{center}
\centerline{\includegraphics[width=\columnwidth]{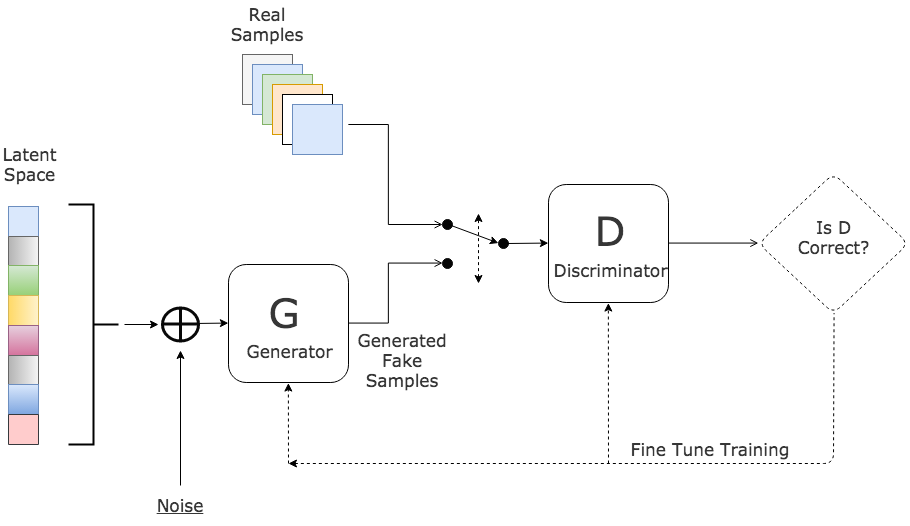}}
\caption{The structure of a Generative Adversarial Network (GAN)}
\label{gan_figure}
\end{center}
\vskip -0.2in
\end{figure}

Here, we have a distribution $p_{data}$ over data $x$ and a prior on input noise variables given by $p_z(z)$. The generator learns through a differentiable function $G(z; \theta_g)$ represented by a multilayer perceptron with parameter $\theta_g$.The discriminator is given by $D(x;\theta_d)$. Also, $D(x)$ gives the probability that $x$ came from data rather than $p_g$. $D$ is trained to maximize $logD(x)$ whereas $G$ is trained to minimize $log(1-D(G(z)))$. The general architecture of a Generative Adversarial Net is shown in Figure 1. 

Goodfellow et al. \yrcite{goodfellow2014generative} proposed an iterative approach of optimization to avoid overfitting the discriminator $D$. Here, they alternate between $k$ steps of optimizing $D$ and one step of optimizing $G$. Also, it was found that equation (1) did not provide sufficient gradient for $G$ to learn well. Thus, the objective function in (1) can be modified such that instead of training $G$ to minimize $log(1-D(G(z)))$, we can train $G$ to maximize $logD(G(z))$. The modified objective function is given as follows:

\begin{multline}
V(D, G) = \max_{D} [E_{x \sim p_{data}(x)}[logD(x)] + \\
E_{z \sim p_{z}(z)}[log(1 - D(G(z)))]] + \max_{G} E_{z \sim p_{z}(z)}[logD(G(z))]
\end{multline}

In the next section, we discuss the various modifications proposed to the Generative Adversarial Networks, their architectures for generative and discriminative models, gradient calculations, experimental performance, usage and their advantages and disadvantages.

\section{Survey}
\label{survey}

\subsection{Conditional Generative Adversarial Networks}

\begin{figure}[ht]
\vskip 0.2in
\begin{center}
\centerline{\includegraphics[width=\columnwidth]{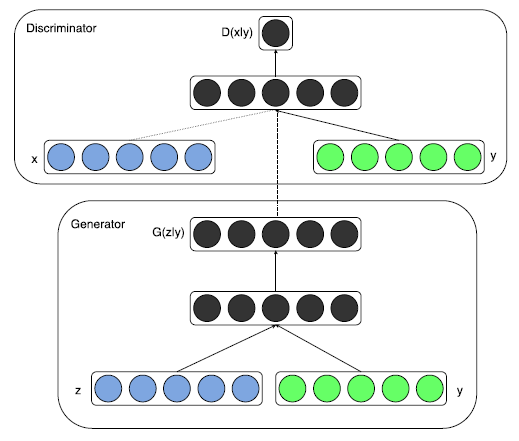}}
\caption{The structure of a Conditional Generative Adversarial Network (CGAN)}
\label{cgan_figure}
\end{center}
\vskip -0.2in
\end{figure}

\begin{figure*}[ht]
\vskip 0.2in
\begin{center}
\centerline{\includegraphics[width=\textwidth]{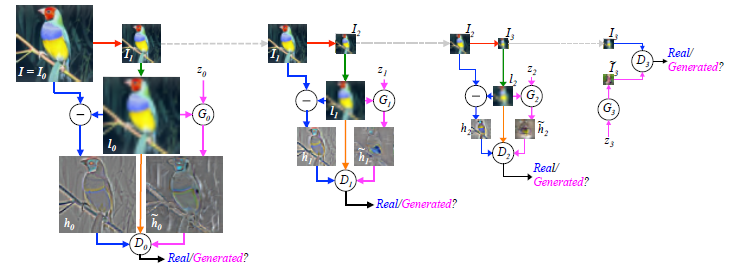}}
\caption{The structure of a Laplacian Pyramid of Generative Adversarial Network (LAPGAN)}
\label{lapgan_figure}
\end{center}
\vskip -0.2in
\end{figure*}

Conditional Generative Adversarial Networks(CGANs) \cite{mirza2014conditional} are the conditional version of GANs which are constructed by feeding the data $y$, we want to condition on the generator and discriminator. Here $y$ can be any kind of information such as class labels or data from other modalities. In the generator for CGANs, the prior input noise $p_z(z)$ and the auxiliary information $y$ are combined in joint hidden representations. In the discriminator, $x$ and $y$ are presented as inputs to the discriminative function. Here, the objective function is similar to that of vanilla GAN except that the data distributions are now conditioned on $y$. This modified objective function is given as follows:

\begin{multline}
\min_{G} \max_{D} V(D, G) = E_{x \sim p_{data}(x)}[logD(x|y)] \\ + E_{z \sim p_{z}(z)}[log(1 - D(G(z|y)))]
\end{multline}

The architectural diagram of CGAN can be seen in Figure 2. Both, the generator and the discriminator are multilayer perceptrons with Rectified Linear Units (ReLU) as the activation for hidden layers and sigmoid for the output layer. The model is trained using stochastic gradient descent with initial learning rate of 0.1 exponentially decreased down to .000001 with a decay factor of 1.00004. Mirza et al. \yrcite{mirza2014conditional} also demonstrate the use of CGANs for automated tagging of images with multi-label predictions. This allows them to generate a distribution of tag-vectors conditional on image features. 

\subsection{Laplacian Pyramid of Adversarial Networks}

Denton et al. \yrcite{denton2015deep} proposed the generation of images in a coarse-to-fine fashion using cascade of convolutional networks within a Laplacian pyramid framework. This approach allowed them to exploit the multiscale structure of natural images, building a series of generative models, each capturing image structure at a particular level of the Laplacian pyramid.

The Laplacian pyramid is built from a Gaussian pyramid using upsampling $u(.)$ and downsampling $d(.)$ functions. Let $G(I) = [I_0, I_1, ..., I_K]$ be the Gaussian pyramid where $I_0 = I$ and $I_K$ is $k$ repeated applications of $d(.)$ to $I$. Then, the coefficient $h_k$ at level $k$ of the Laplacian pyramid is given by the difference between the adjacent levels in Gaussian pyramid, upsampling the smaller one with $u(.)$. 

\begin{multline}
h_k = L_k(I) = G_k(I) - u(G_{k+1}(I) = I_k - u(I_{k+1}))
\end{multline}

Reconstruction of the Laplacian pyramid coefficients $[h_1, ..., h_K]$ can be performed through backward recurrence as follows:

\begin{equation}
I_k = u(I_{k+1} + h_k)
\end{equation}

Thus, while training a LAPGAN, we have a set of convolutional generative models ${G_0, ..., G_K}$, each of which captures the distribution of coefficients $h_k$ for different levels of the Laplacian pyramid. Here, while reconstruction, the generative models are used to produce $h_k$'s. Thus, equation (5) gets modified as follows:

\begin{multline}
\widetilde{I}_k = u(\widetilde{I}_{k+1}) + \widetilde{h}_k = u(\widetilde{I}_{k+1}) + G_k(z_k, u(\widetilde{I}_{k+1}))
\end{multline}

Here, a Laplacian pyramid is constructed from each training image $I$. At each level a stochastic choice is made regarding constructing the coefficient $h_k$ using the standard procedure or generate them using $G_k$. The entire procedure for training a LAPGAN through various stages can be seen in Figure 3.

LAPGANs also take advantage of the CGAN model by adding a low-pass image $l_k$ to the generator as well as the discriminator. The authors evaluated the performance of the LAPGAN model on three datasets: (i)CIFAR10 (ii)STL10 and (iii) LSUN datasets. This evaluation was done by comparing the log-likelihood, quality of image samples generated and a human evaluation of the samples.

\subsection{Deep Convolutional Generative Adversarial Networks}

Radford et al. \yrcite{radford2015unsupervised} proposed a new class of CNNs called Deep Convolutional Generative Adversarial Networks (DCGANs) having certain architectural constraints. These constraints involved adopting and modifying three changes to the CNN architectures. 

\begin{itemize}
\item Removing fully-connected hidden layers and replacing the pooling layers with strided convolutions on the discriminator and fractional-strided convolutions on the generator
\item Using batchnormalization on both the generative and discriminative models\
\item Using ReLU activations in every layer of the generative model except the last layer and LeakyReLU activations in all layers of the discriminative model
\end{itemize}

\begin{figure}[ht]
\vskip 0.2in
\begin{center}
\centerline{\includegraphics[width=\columnwidth]{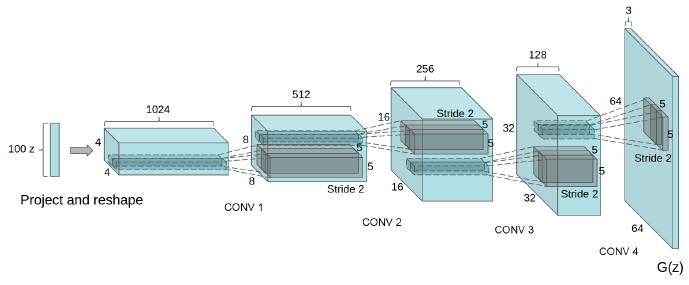}}
\caption{The structure of a Deep Convolutional Generative Adversarial Network (DCGAN)}
\label{dcgan_figure}
\end{center}
\vskip -0.2in
\end{figure}

Figure 4 depicts the DCGAN generator for LSUN sample scene modeling. The DCGAN model's performance was evaluated against LSUN, Imagenet1k, CIFAR10 and SVHN datasets. The quality of unsupervised representation learning was evaluated by first using DCGAN as a feature extractor and then the performance accuracy was calculated by fitting a linear model on top of those features. Log-likelihood metrics were not used for performance evaluation. The authors also demonstrated feature learning by the generator showcasing how the generator could learn to forget scene components such as bed, windows, lamps and other furniture. They also performed vector arithmetic on face samples leading to good results.

\subsection{Adversarial Autoencoders}

Makhzani et al. \yrcite{makhzani2015adversarial} proposed adversarial autoencoder which is a probabilistic autoencoder which makes use of GAN to perform variational inference by matching the aggregated posterior of the hidden code vector  of the autoencoder with an arbitrary prior distribution. In adversarial autoencoder, the autoencoder is trained with dual objectives - a traditional reconstruction error criteria, and an adversarial training criterion that matches the aggregated posterior distribution of the latent representation to an arbitrary prior distribution. After training, the encoder learns to convert the data distribution to the prior distribution, while the decoder learns a deep generative model that maps the imposed prior to the data distribution. The architectural diagram of an adversarial autoencoder is shown in Figure 5.

\begin{figure}[ht]
\vskip 0.2in
\begin{center}
\centerline{\includegraphics[width=\columnwidth]{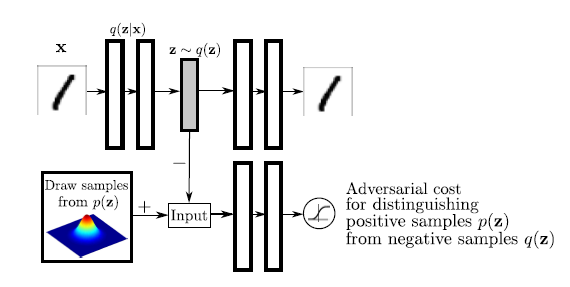}}
\caption{The structure of an Adversarial Autoencoder (AAE)}
\label{aae_figure}
\end{center}
\vskip -0.2in
\end{figure}

Let $x$ be the input and $z$ be the latent code vector of an autoencoder. Let $p(z)$ be the prior distribution we want to impose, $q(z|x)$ be the encoding distribution and $p(x|z)$ be the decoding distribution. Also, let $p_d(x)$ be the data distribution and $p(x)$ be the model distribution. The encoding function of the autoencoder $q(z|x)$ defines an aggregated posterior distribution of $q(z)$ on the hidden code vector of the autoencoder as follows:

\begin{equation}
q(z) = \int_x q(z|x)p_d(x)dx
\end{equation}

In adversarial autoencoder, the autoencoder is regularized by matching the aggregated posterior $q(z)$ to an arbitrary prior $p(z)$. The generator of the adversarial network is also the encoder of the autoencoder $q(z|x)$. Both, the adversarial network and the autoencoder are trained jointly with stochastic gradient descent in two phases - the reconstruction phase and the regularization phase. In the reconstruction phase, the autoencoder updates the encoder and the decoder to minimize the reconstruction error of the inputs. In the regularization phase, the adversarial network first updates the discriminator to tell apart the true samples from the generated ones and then updates the generative model in order to confuse the discriminator. 

Labels can also be incorporated in AAEs in the adversarial training phase in order to better shape distribution of the hidden code. A one-hot vector is added to the input of the discriminative network to associate the label with the mode of distribution. Here, the one-hot vector acts as a switch that selects the corresponding decision boundary in the discriminative network given the class label. The one-hot vector also contains one point corresponding to an extra class which in turn corresponds to unlabelled examples. When an unlabelled example is encountered, the extra class is turned on and the decision boundary  for the full mixture of Gaussian distribution is selected.

The authors evaluated the performance of adversarial autoencoders on MNIST and Toronto Face datasets using log-likelihood analysis in supervised, semi-supervised and unsupervised settings. In supervised settings, the one-hot vector of class labels is encoded and provided to the decoder. The decoder utilizes both the one-hot vector and the hidden code $z$ for reconstructing the image.In semi-supervised settings, the generative description of unlabelled data is exploited. Here, it is assumed that data is generated by a latent class variable $y$ that comes from a categorical distribution as well as a continuous latent variable $z$ that comes from a Gaussian distribution.

\begin{equation}
p(y) = Cat(y) \qquad p(z) = N(z|0, I)
\end{equation}

Here, the adversarial network and the autoencoder are trained in three phases viz. the reconstruction phase, regularization phase and the semi-supervised classification phase. The first two phases are similar to those in supervised settings. In the semi-supervised classification phase the autoencoder updates $q(y|x)$ to minimize the cross-entropy cost. In unsupervised settings, clustering is done using a model with an architecture similar to that for semi-supervised settings except for the fact that there is no semi-supervised classification stage. Moreover, the inference network $q(y|x)$ predicts a one-hot vector whose dimension is the number of categories that the data can be clustered into. The authors also show how adversarial autoencoders can be used for dimensionality reduction.

\subsection{Generative Recurrent Adversarial Networks}
Im et al. \yrcite{im2016generating} proposed recurrent generative model showing that unrolling the gradient based optimization yields a recurrent computation that creates images by incrementally adding to a visual ``canvas". Here, the ``encoder" convolutional network extracts images of current ``canvas". The resulting code and the code for the reference image get fed to a ``decoder" which decides on an update to the ``canvas". Figure 6 depicts an abstraction of how a Generative Recurrent Adversarial Network works. The function $f$ serves as the decoder and the function $g $serves as encoder in GRAN.

\begin{figure}[ht]
\vskip 0.2in
\begin{center}
\centerline{\includegraphics[width=\columnwidth]{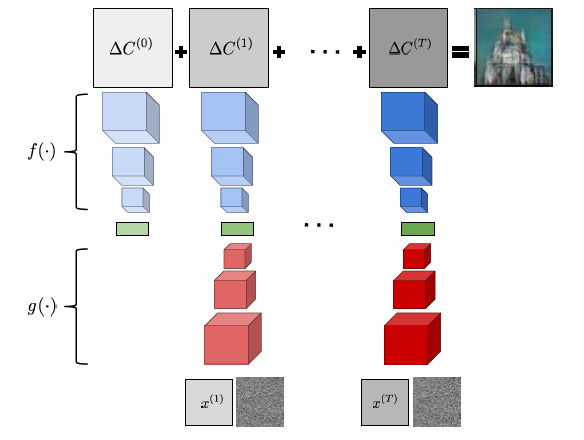}}
\caption{The structure of a Generative Recurrent Adversarial Network (GRAN)}
\label{gran_figure}
\end{center}
\vskip -0.2in
\end{figure}

In GRAN, the generator $G$ consists of  a recurrent feedback loop that takes a sequence of noise samples dawn from the prior distribution $z \sim p(z)$ and draws the output at different time steps $C_1, C_2, ..., C_T$. At each time step $t$, a sample $z$ from the prior distribution is passed onto a function $f(.)$ with the hidden state $h_{c, t}$ where $h_{c, t}$ represents the current encoded status of the previous drawing $C_{t-1}$. $C_t$ is what is drawn to the canvas at time $t$ and it contains the output of the function $f(.)$. Moreover, the function $g(.)$ is used to mimic the inverse of the function $f(.)$. Accumulating the samples at each time step yields the final sample drawn to the canvas $C$. Ultimately, the function $f(.)$ acts as a decoder and receives the input from the previous hidden state $h_{c, t}$ and noise sample $z$ and the function $g(.)$ acts as an encoder that provides a hidden representation of the output $C_{t-1}$ for time step $t$. Interestingly, compared to all other auto-encoders which start by encoding an image, GRAN starts with a decoder.

The authors propose a new evaluation metric for generative models called Generative Adversarial Metric (GAM). This metric compares two generative adversarial models by having them engage in a ``battle" against each other. Here the generator and discriminator of the models are exchanged and pitted against each other. If $M_1$ and $M_2$ are two generative adversarial models, then after training these models during the test phase, model $M_1$ plays against model $M_2$ by having $G_1$ trying to fool $D_2$ and vice versa. Also, two ratios using the discriminative scores of these models are calculated:

\begin{equation}
r_{test} \myeq \frac{\epsilon(D_1(x_{test}))}{\epsilon(D_2(x_{test}))}
\end{equation}
\begin{equation}
r_{samples} \myeq \frac{\epsilon(D_1(G_1(z)))}{\epsilon(D_2(G_2(z)))}
\end{equation}

where $\epsilon(.)$ outputs the classification error rate. The test ratio, $r_{test}$, tells us which model generalizes better as it is based on discriminating the test data. The sample ratio, $r_{samples}$, tells us which model can fool the other model more easily, since the discriminators are classifying over the samples generated by their opponents. The proposed evaluation metrics qualify the sample ratio using the test ratio by defining the winning model as follows:

\begin{equation}
winner = 
\begin{cases}
M1 \quad if \> r_{sample} < 1 \quad and \quad r_{test} \simeq 1 \\ 
M2 \quad if \> r_{sample} > 1 \quad and \quad r_{test} \simeq 1 \\
Tie \quad otherwise
\end{cases}
\end{equation}
 
The GRAN model's performance was evaluated against MNIST, CIFAR10 and LSUN datasets with time steps $T = {1, 3, 5}$. It was found that GRAN with time steps $T = 3$ and $T = 5$ performed better than GRAN with time step $T = 1$. Also, GRAN was compared against other generative models such as denoising VAE and DRAW on the MNIST dataset. It was also found that the samples generated by GRAN were discernible and did not overfit on the training data.

\subsection{Information Maximizing Generative Adversarial Networks}
Information maximizing GANs (InfoGANs) \cite{chen2016infogan} are an information-theoretic extension of GANs that are able to learn disentangled features in a completely unsupervised manner. A disentangled representation is one which explicitly represents the salient features of a data instance and can be useful for tasks such as  face recognition and object recognition. Here, InfoGANs modify the objective of GANs to learn meaningful representations by maximizing the mutual information between a fixed small subset of GAN's noise variables and observations.

In GANs, there are no restrictions on the manner in which the generator may use the noise. As a result, the noise may be used in a highly entangled way not corresponding to the semantic features of the data. However, it makes sense to semantically decompose a domain according to the semantic features of the data under consideration. InfoGANs use this approach by decomposing the input noise vector into two parts: (i)$z$ which is treated as a source of noise, (ii) $c$ called the latent code and targeted at the salient structured semantic features of the data distribution. Thus, the generator network with both the incompressible noise $z$ and the latent code $c$ becomes the generator $G(z, c)$. In order to avoid the latent code $c$ being ignored, information-theoretic regularization is done and the information $I(c; G(z, c))$ is maximized. The information regularized minimax game is given as follows:

\begin{multline}
\min_G \max_D V_1(D, G) = V(D, G) - \lambda I(c; G(z, c))
\end{multline}

Experiments were performed on MNIST, SVHN, CelebA and chairs datasets. Disentangled representations could easily be learned through the usage of discrete and continuous latent codes. For MNIST dataset, one categorical code was used to model discontinuous variation in data and two continuous codes were used to capture style such as rotation and width of digits. Similarly, for the faces dataset, a disentangled representation of azimuth(pose), elevation and lighting were captured using continuous latent variables. Thus, InfoGANs can perform disentanglement high semantic variations such as the presence or absence of sunglasses, hairstyles and emotions without any supervision. Additionally, they require negligible computation cost on top of GANs and are easy to train. 

\subsection{Bidirectional Generative Adversarial Networks}

\begin{table*}[t]
\caption{Comparison of different versions of GANs based on various criteria.}
\label{comparison-table}
\vskip 0.15in
\begin{center}
\begin{large}
\resizebox{\textwidth}{!}{\begin{tabular}{|C{3cm}|C{3cm}|C{3cm}|C{3cm}|C{3cm}|C{3cm}|C{3cm}|C{3cm}|C{3cm}|}
\hline
\abovespace\belowspace
\textsc{Criteria} & \textsc{Vanilla GAN} & \textsc{CGAN} & \textsc{LAPGAN} & \textsc{DCGAN} & \textsc{AAE} & \textsc{GRAN} & \textsc{InfoGAN} & \textsc{BiGAN} \\
\hline
\abovespace
Learning    			& Supervised & Supervised & Unsupervised & Unsupervised & Supervised, semi-supervised and unsupervised & Supervised & Unsupervised & Supervised and unsupervised \\
\hline
Network Architecture    & Multilayer perceptrons & Multilayer perceptrons & Laplacian pyramid of convolutional networks & Convolutional networks with constraints & Autoencoders & Recurrent convolutional networks with constraints & Multilayer perceptrons & Deep multilayer neural networks \\
\hline
Gradient Updates        & SGD with k steps for D and 1 step for G & SGD with k steps for D and 1 step for G & No updates & SGD with Adam optimizer for both G and D & SGD with reconstruction and regularization steps & SGD updates to both G and D & SGD updates to both G and D & No updates \\
\hline
Methodology / Objective & Minimize value function for G and maximize for D & Minimize value function for G and maximize for D conditioned on extra information & Generation of images in coarse-to-fine fashion & Learn hierarchy of representations from object parts to scenes in both G and D & Inference by matching posterior of hidden code vector of autoencoder with prior distribution & Generation of images by incremental updates to a ``canvas" & Learn disentangled representations by maximizing mutual information & Learn features for related semantic tasks and use in unsupervised settings \\
\hline
Performance Metrics     & Log-likelihood & Log-likelihood & Log-likelihood and human evaluation & Accuracy and error rate & Log-likelihood and error-rate & Generative Adversarial Metric (proposed) & Information metric and representation learning & Accuracy \\
\hline
\end{tabular}}
\end{large}
\end{center}
\vskip -0.1in
\end{table*}

Donahue et al. \yrcite{donahue2016adversarial} proposed a method for learning the semantics in data distribution as well as its inverse mapping - using these learnt feature representations for projecting data back into the latent space. The structure of a Bidirectional Generative Adversarial Network is shown in Figure 7. As it can be seen from the figure, in addition to the generator $G$ from the standard GAN framework, BiGAN includes an encoder $E$ which maps the data $x$ to latent representations $z$. The BiGAN discriminator $D$ discriminates not only in the data space ($x$ versus $G(z)$), but jointly in data and latent spaces (tuples $(x, E(x))$ versus $(G(z), z)$), where the latent component is either the encoder output $E(x)$ or generator input $z$. Here, according to the objective of GANs, the BiGAN encoder $E$ should learn to invert the generator $G$. The BiGAN training objective is defined as follows:

\begin{multline}
\min_{G, E} \max_D V(D, E, G) = E_{x \sim p_x} \underbrace {E_{z \sim p_E(.|x)}[logD(x, z)]}_{logD(x, E(x))} \\ +  
E_{z \sim p_z} \underbrace{E_{x \sim p_G(.|z)}[1-logD(x, z)]}_{log(1-D(G(z), z))}
\end{multline}

\begin{figure}[ht]
\vskip 0.2in
\begin{center}
\centerline{\includegraphics[width=\columnwidth]{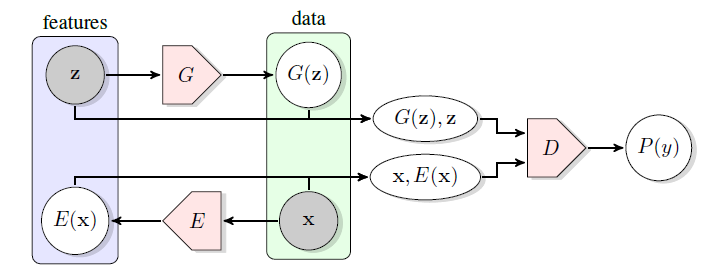}}
\caption{The structure of a Bidirectional Generative Adversarial Network (BiGAN)}
\label{bigan_figure}
\end{center}
\vskip -0.2in
\end{figure}

BiGAN share many properties of GANs while additionally guaranteeing that at the global optimum, $G$ and $E$ are each other's inverse. Learning in BiGAN is done using stochastic gradient descent by learning parameters $\theta_D$, $\theta_G$ and $\theta_E$ for modules $D$, $G$ and $E$. In an iteration of optimization using stochastic gradient descent, the discriminator parameters $\theta_D$ are updated by taking one or more steps in the positive gradient direction. Then, the encoder and generator parameters, $\theta_E$ and $\theta_G$ are updated by taking one step in the negative gradient direction. It was also observed that training an inverse objective for $E$ and $G$ provided stronger gradients. 

The authors evaluated their BiGAN model by first training them unsupervised and then transferring the encoder's learned feature representations for supervised learning tasks. For permutation invariant MNIST dataset, each module was trained as a multilayer perceptron and performed comparatively at the same level as other methods. Next, BiGAN were trained on the ImageNet LSRVC dataset where each module is a convolutional network. It was found that the convolutional filters learned by the encoder $E$ had a Gabor-like structure. Also, the BiGAN encoder $E$ and the generator $G$ learned approximate inverse mappings.

\section{Analysis}
\label{analysis}

This section discusses different versions of GANs and provides a comparison between them. 

It can be seen from Table 1. that initial versions of GANs such as Vanilla GAN and Conditional GAN only supported supervised learning which were later augmented to support semi-supervised and unsupervised learning. Moreover, earlier adversarial frameworks used multilayer perceptrons which were later experimented with other network structures such as convolutional networks, autoencoders and deep neural networks. Also, for most of the models, Stochastic Gradient Descent based optimization was used for training both the generator and discriminator networks.

The primary objective of any adversarial network remains a 2-player minimax game over all versions. Additionally, some models had secondary objectives such as feature learning and learning of representations through related semantic tasks and then later using these learned features for classification or recognition in unsupervised settings. Also, models such as LAPGAN and GRAN introduced a sequential generation of images by the generator using Laplacian pyramids and recurrent networks. 

Additionally, earlier models evaluated model performance on the basis of log-likelihood estimates which was discarded in later versions as it was not a good estimate. Instead accuracy and error rates were used for evaluating the performance of a model. Also, GRAN proposed a new evaluation metric called Generative Adversarial Metric for evaluating the performance of generative adversarial nets although it has not been in use by any other generative model.

\section{Conclusion and Future Work}
\label{conclusion}

This paper provides a much-needed comparative analysis of different versions of Generative Adversarial Networks on the basis of their methodology, architecture and performance. It can be seen that the later versions of adversarial networks are more robust and have many more applications compared to the original version. Also, these networks can prove to be useful in image classification, recognition, capturing and generation in a variety of ways.

This work can be extended by comparing more recent versions of GANs such as Boundary-Seeking Generative Adversarial Networks \cite{hjelm2017boundary}, Wasserstein Generative Adversarial Networks (WGAN) \cite{arjovsky2017wasserstein}, Loss-Sensitive Generative Adversarial Networks \cite{qi2017loss}, Speech Enhancement Generative Adversarial Networks (SEGAN) \cite{pascual2017segan} and Layered-Recursive Generative Adversarial Networks (LR-GAN) \cite{yang2017lr}. Also, an empirical analysis can be performed on all these versions using benchmark datasets such as MNIST, CIFAR10 and ImageNet. Finally, a library of these models could be implemented allowing future researchers to utilize these models to their fullest in various applications related to images as well as allowing them to replicate results. 

\section*{Acknowledgements}
\label{acknowledgements}

The author would like to thank Prof. Kate Larson for providing useful insights. Her comments and feedback were helpful in modelling and conducting the review.

\bibliography{report}
\bibliographystyle{icml2015}

\end{document}